\documentclass{article}
\usepackage{spconf,amsmath,graphicx}
\usepackage{comment}
\usepackage{float} 

\title{ADAPTIVE LARGE LANGUAGE MODELS BY LAYERWISE ATTENTION SHORTCUTS}
\name{Prateek Verma and Mert Pilanci}
\address{Department of Electrical Engineering \\ Stanford University\\
Stanford, CA, 94305\\}
\newcommand\blfootnote[1]{%
  \begingroup
  \renewcommand\thefootnote{}\footnote{#1}%
  \addtocounter{footnote}{-1}%
  \endgroup
}
\begin{document}
%
\maketitle
\begin{abstract}
Transformer architectures are the backbone of the modern AI revolution. However, they are based on simply stacking the same blocks in dozens of layers and processing information sequentially from one block to another. In this paper, we propose to challenge this and introduce adaptive computations for LLM-like setups, which allow the final layer to attend to all of the intermediate layers as it deems fit through the attention mechanism, thereby introducing computational \textbf{attention shortcuts}. These shortcuts can thus make the architecture depth and context adaptive. We showcase four different datasets, namely acoustic tokens, natural language, and symbolic music, and we achieve superior performance for GPT-like architecture. We give evidence via attention maps that the models learn complex dependencies across layers that are adaptive in context and depth depending on the input tokens. \blfootnote{This work was supported in part by the National Science Foundation (NSF) under Grant DMS-2134248; in part by the NSF CAREER Award under Grant CCF-2236829; in part by the U.S. Army Research Office Early Career Award under Grant W911NF-21-1-0242; and in part by the Office of Naval Research under Grant N00014-24-1-2164.}
\end{abstract}
\section{Introduction}
\label{sec:intro}
Transformer architectures have revolutionized modern AI advancements, powering almost all diverse fields with a unified approach \cite{vaswani2017attention}. Modern Transformer language models consist of a simple stack of Transformer decoder blocks iteratively in modalities such as text \cite{brown2020language}, raw audio waveforms\cite{verma2021generative}, acoustic and music tokens \cite{huang2018music, verma2020framework,borsos2023audiolm}, videos \cite{yan2021videogpt} to name a few. The flexibility of the architecture to handle any input has generated an enormous interest in the field, and the advent of multi-modal architectures proposed by Google with its Gemini family \cite{team2023gemini} or multi-modal models like \cite{Chameleon} that can allow these models to hear, see and read. However, the information is processed sequentially in all these architectures, one block after another, to allow features to learn representations at multiple levels and depths. In this paper, we explore if we can make these architectures' depth and context adaptive depending on the contents of the input token. We allow the final layer's attention mechanism to bypass intermediate layers' computation and directly contribute to the next token prediction. This would allow more straightforward tokens present in the input, which are easier to predict, to directly learn features in shallow layers
to predict the outout. It can thus better utilize and reserve deeper complex self-attention blocks for tougher token predictions. Mainly two ideas inspired this paper -- First, several papers utilize intermediate layers that capture information at various scales \cite{hewitt2019structural,tenney2019bert,song2020utilizing, oh2022don}, for solving downstream tasks in natural language such as sentiment analysis, word representations, etc. These intermediate representations are often helpful for capturing useful features of text at multiple scales.
Further, such behavior is observed for modalities like speech, where \cite{pasad2023comparative} showed it  that different layer representations were useful for transformer-based architecture for several downstream tasks related to speech recognition and spoken language understanding. Even for a seq-to-seq architecture, Whisper-AT \cite{gong2023whisper} showed the importance of intermediate layers by using a frozen Whisper-encoder and a temporal transformer to learn feature mapping to classify acoustic sounds. With all of these, we want to state that there is enough evidence to show that intermediate Transformer layers capture meaningful information, capturing different aspects of input signals across modalities. In this paper, we answer --

\vspace{0.02cm}
\textbf{1. Can we leverage intermediate layer embeddings for improving pre-training of a Large Langauge Model}? 
\vspace{0.02cm}

The second evidence we address in our architecture is that for certain kinds of inputs, the output of several Transformer layers can be approximated by a single MLP layer, as shown in a recent paper \cite{hernandez2023linearity}. This is fascinating find as they showed that for specific input tokens, we can bypass mapping of several layers in depth and context to predict the same output as a simple function approximation. Our paper allows this behaviour to be learned during pre-training. This allows our model to avoid wasting complex attention layers when learning trivial feature maps. This is the basis of our paper  -- 

\begin{figure}[t!] 
  \centering
  \hspace*{-8.8pt}
  \includegraphics[width=0.9\linewidth,height=5cm]{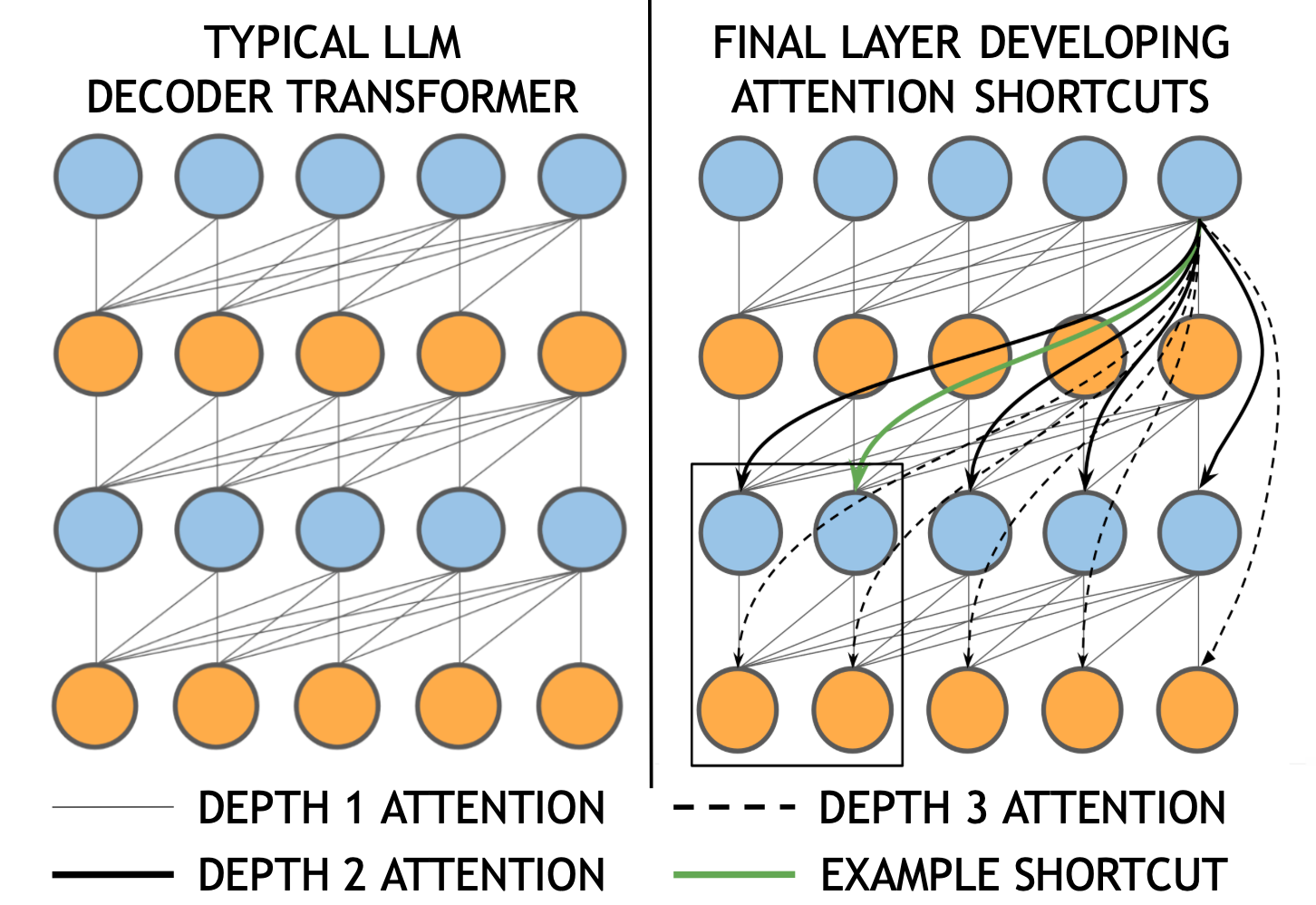} 
  \caption{(Left) Typical Transformer LLM (Middle) Our proposed Adaptive LLM that attends to intermediate layer embedding, allowing us to learn adaptive shortcuts both in context and depth. When predicting the last token, model's final layer can now attend to embeddings(and derived features) at different depths and contexts as it deems fit. The dark curved connectors attend to the second layer, and the dotted connectors attend to the first layer while decoding from the fourth layer. E.g. for the last token in the final layer, it can develop a connection in green to learn a shortcut directly from the 2nd layer second token via attention (Right). Our network is dense with more connections allowed than the vanilla Transformer. }
  \label{fig:adaptive_LLM}
\end{figure}

\vspace{0.02cm}
\textbf{2. Can we make LLM context and depth adaptive with input content to bypass several layers and tokens in context by learning simple feature maps for the final layer? }
\vspace{0.02cm}This differs from Pathway's architecture \cite{chowdhery2023palm} proposed by Google (Palm) as they have separate streams of MLP and attention blocks, and the overall output is the addition of the output of two blocks. However, the attention mechanism is still confined to the current layer depth. It is not adaptive to allow for a direct bypass of several layers to the output in context and depth. 
The contributions of the paper are as follows: 

\begin{itemize}
   \item To the best of our knowledge, we, for the first time, show the powerfulness of a decoder-only transformer architecture to attend to all intermediate layer features to improve pre-training performance.
    \item We show for four different datasets that our method improves the pre-training performance and gives evidence via attention maps that model learns \textbf{attention shortcuts} by attending to the desired depth, depending on the complexity of the current token and its past context.
\end{itemize}
\label{sec:intro}
\vspace{-0.2cm}
\section{Dataset}
\label{sec:dataset} We utilize four publicly available commonly used datasets for showcasing the powerfulness of our method, with context length of 512. These are chosen from three modalities namely speech, symbolic music and natural language. For speech, we take LibriSpeech corpus \cite{panayotov2015librispeech}, which consists of 960 hours of spoken speech for training. We convert it into discrete tokens using ENCODEC \cite{defossez2022high}. We use the entire training corpus i.e. train-clean-100/360, and train-other-500 as our training data. This gave us a total of 270 million tokens.  We utilize the coarsest codebook as proposed in Vall-E \cite{wang2023neural} to report our results. The model is trained with the vocabulary size of 1024 tokens. For natural language we used two dataset, using two different tokenizer. The first is character level language modelling on text-8 which uses 26 characters and an extra space token as its vocabulary with all other text or symbols removed and replaced with space. This choice removes any biases that may occur due to tokenizers and the dataset itself is widely reported. It has 100M characters from wikipedia. We also use Wiki-103 dataset with GPT-2 tokenzier that has 50,257 tokens. For both of them, we randomly crop 1 million crops with 512 context length yielding a total of 512 million tokens. This is done to have variable context for the same token to make the data more robust. Finally, we use MAESTRO \cite{hawthorne2018enabling}, which consists of 1000 piano pieces from classical music in MIDI format. We convert the MIDI track to discrete tokens using Google's tokenizer yielding 388 sized vocabulary. We compare the performance of our architecture with and without the modifications proposed. The paper is written in an academic setting on shrunk down version of GPT architecture ubiquitously used for language modelling. 

\section{Methodology} In all methods, we use Transformer decoder layers. The architecture for all modalities is the same except for attention mechanism of the final layer. We have already mentioned vocabulary sizes for different modalities in the dataset section. For all the experiments, as a baseline, we use a stack of 10 Transformer decoder layers with a context length of 512, a model dimension of 128, and eight attention heads. The size of the feed-forward dimension was chosen to be four times that of the model dimension. The models were trained for 12-13 epochs, thus seeing more than 10 billion tokens. All architectures were trained with Adam optimizer with starting learning rate 2e-4 and then divide by half mid-way. We take the same architecture with the following modifications for our proposed architecture. The output of each Transformer layer in the second, fourth, and eighth layers is passed to a feature learning module, a 2-layer MLP with 1024 neurons, followed by a dense layer of the size of the model dimension. This allows us to learn feature maps out of embeddings at intermediate layers from depths for all the tokens present in our context. A standard decoder only Transformer formulation is,
$$
x^{l+1} = x^{l} + \text{MLP}(\text{LayerNorm}(x^{l} + \text{Attention}(\text{LayerNorm}(x^{l}))))
$$
where $x^{l+1}$, is the output of the $l^{th}$ Transformer decoder layer. Our paper introduces an MLP layer where we learn features directly from $x^{l}$, and we call it $x_{feat}^{l}$ which is, $x_{feat}^{l} = \text{MLP}(x^{l})$. Now, we keep the Transformer decoder blocks similar to the baseline architecture except for the final layer decoder block where we allow features with \textbf{Attention shortcuts}: \textit{We allow the final prediction at every token to bypass intermediate computations both in depth and context as it deems fit}. It allows to directly attend to the desired feature of interest depending on complexity of input tokens and its past context. Further, as shown in the Figure 1, we want the computations to be adaptive, i.e., can we bypass in-depth and context? This follows \cite{hernandez2023linearity}, which found that a set of layers and context can be bypassed with simple features for easier input tokens and context. We focused on addressing these two insights when designing our architecture. In the ablation studies discussed in Section 4.2, we experimented with i) using at a particular depth $l$, learned features with MLP layer $x_{feat}^{l}$ ii) Simply using the $x^{l}$ to learn attention shortcuts. How can we learn computation shortcuts? Since most of the Transformer based LLMs  are simply a stack of Transformer Decoder layers, we allow the architecture to use the intermediate layers. In addition to learning computational attention shortcuts, the other added benefit is that, as we have seen in the introduction, the intermediate layer has such rich information about the input data, which might be more useful than just learning features in a sequential hierarchy. Further, by giving the network the flexibility to attend to the intermediate layer feature maps directly, we do not waste precious resources which are powerful attention blocks and layers just learning simple linear maps as seen in \cite{hernandez2023linearity}. \begin{figure}[] 
  \centering
  \hspace*{-10.8pt}
  \includegraphics[width=\linewidth,height=6.3cm]{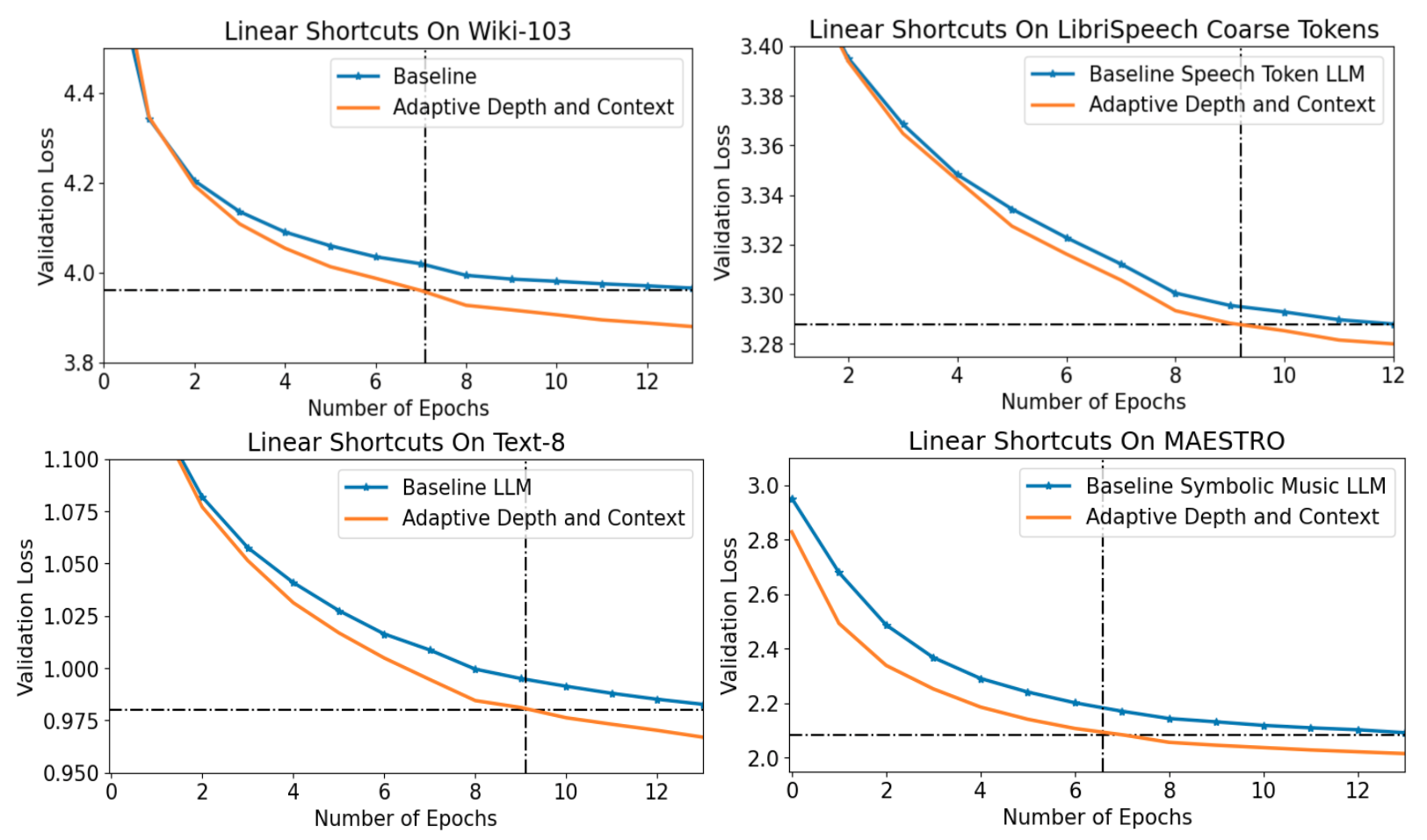}
  \caption{Results of four different datasets for symbolic music, speech tokens and natural langauge.}
  \label{fig:adaptive_LLM}
\end{figure}By allowing the final decoder layer to directly attend to the intermediate features of the tokens at various depths and contexts, as we can see in Figure 1, we address the points that we have described above. We replace the self-attention block of the last layer of the stack of the Transformer decoder with a cross-attention layer. The attention mask has to be causal. Hence we only allow for the final Transformer block to attend to all of the previous current layer inputs $x^{l}(t)$, ($l$ being 9 in our case), and all features $x_{feat}^{2}(k)$, $x_{feat}^{4}(k)$, $x_{feat}^{6}(k)$, $x_{feat}^{8}(k)$ with $k$ going from $1,2,3, t$, and $L$ being context length as 512. We introduce causal masks so that at any instance $t\le L$, $x^{9}(t)$ can attend to all features derived from intermediate layers  $x_{feat}^{l}(k)$ i.e. $x_{feat}^{2}(k)$, $x_{feat}^{4}(k)$, $x_{feat}^{6}(k)$, $x_{feat}^{8}(k)$, with $k$ going from $1,2,3,...t$. This formulation of cross attention is similar to perceiver AR \cite{hawthorne2022general}, Figure 1, using it for LLM like setups. We learn the value and the key matrices are learned from the concatenated stack of $x_{feat}^{2}(k)$, $x_{feat}^{4}(k)$, $x_{feat}^{6}(k)$, $x_{feat}^{8}(k)$, and the value matrices are learned from $x^{l}(k)$, with proper masking to ensure we do not attend to the future tokens. Since there are now 4 times more tokens to attend to in the cross attention layer, we have twice the number of heads just in the final cross attention layer to account for that. Our model has the same structure in all the layers as a vanilla transformer architecture except for using cross-attention in the final layer before softmax classification instead of a self-attention layer. Thus, the final layer decoder output $x^{10}$ with stacked features $x_{feat} = [x_{feat}^{2}, x_{feat}^{4}, x_{feat}^{6}, x_{feat}^{8}], $ is
\vspace{-0.2cm}
$$
\begin{aligned}
x^{10} &= x^{9} + \text{MLP}(\text{LayerNorm}(x^{9} + x_{att})) \\
x_{att} &= \text{CrossAttention}(\text{LayerNorm}(x^{9}), \text{LayerNorm}(x_{feat})) \\
\end{aligned}
$$
\vspace{-0.8cm}
\section{Results and Discussion}
We wanted to showcase the power of our architecture across four different datasets and modalities. As described in the previous section, we use text-8, Wiki-103, LibriSpeech, and MAESTRO. We compare the performance of our architecture with that of the baseline only in terms of negative log-likelihood scores. This is in line with other papers in natural language processing \cite{al2019character,yu2024megabyte} that report the performance of architecture only in terms of advancement of how well it does in pretraining in likelihood scores. Neural architectures scale well with the size of the architecture \cite{kaplan2020scaling}. In addition, keeping all the aspects of the training recipe the same, i.e., training corpus, optimizers, and post-training like DPO or RLHF, the reason for the improved performance of LLMs is their ability to predict the next token correctly. Large Language Model-based generative architectures utilize sampling from the distribution of the next token to solve generative tasks, e.g., music generation, text generation, etc. Large architecture performs better because fewer errors in predicting the next token are present due to the probability distribution closer to the desired distribution. Further, the importance of likelihood-based metrics can also judged by the fact that in latest advanced models, like Gemini, these scores were conspicuously absent and were left blank from the results and plots. Hence, with this background, we will begin discussing our results. 
\vspace{-0.5cm}
\begin{table}[ht]
  \caption{Comparison of the negative-log likelihood (NLL) scores (log base e) for our architecture for different datasets with/without adding layer-wise attention on feature maps}
	\centering
	\begin{tabular}{|c|c|c|c|c|}
		\hline
		Dataset & Baseline & Ours & SP Epoch & \% Speed-Up \\\hline
        Text-8  & 0.99  & 0.97 & 9.1/13 & 30\% \\
		Wiki-103 &3.96  & 3.85 & 6.8 /13 & 47.7\% \\
        \textbf{MAESTRO}  & \textbf{2.10}  & \textbf{2.02} & \textbf{6.6 /13} & \textbf{49.2\%} \\
        LibriSpeech  & 3.28  & 3.27 & 9.2 /12 &30.4\%\\\hline
		
	\end{tabular}
	\label{tab:example}
\end{table}

\vspace{-0.01cm}
\subsection{Performance across modalities} For all four datasets mentioned in Section 2, we perform our experimentation as described in Table 1. We can see that our methodology is generic across four datasets. We report how well we do in terms of next token prediction and report the negative log-likelihood loss. We can see that we achieve the best speed-up for the MAESTRO corpus; one reason is that music has associated rules, and the lower layers can quickly learn them while saving complex dependencies for the later layers. We compare the performance with the baseline architecture, which did not have a final attention layer attending to intermediate layers embeddings at various depths. Speed-up (SP), which is the time to get the same validation loss as baseline model by our proposed method is also reported. 
\vspace{-0.5cm}
\begin{table}[ht]
  \caption{Varation for configurations}
	\centering
	\begin{tabular}{|c|c|c|}
		\hline
		Feature Set For Wiki-103 & NLL & Accuracy\\\hline
		Baseline  &  3.95 & 31.78\% \\
        Middle Layer with MLP Features & 3.92 & 32.24\%\\
       \textbf{All Layers -- with MLP Features} & \textbf{3.86} & \textbf{32.81\%} \\
       All Layers -- without MLP Features &  3.93 & 32.02\% \\
       \hline		
	\end{tabular}
	\label{tab:example}
\end{table}
\vspace{-0.7cm}
\subsection{Ablation Studies} For Wiki-103, we carry out ablation studies on our proposed architecture as compared to a baseline model to see the effects of i) allowing the final layer Transformer decoder to attend to all of the intermediate representation and ii) only allowed to attend to intermediate representation in the middle layers iii) Seeing the effects of without/without adding MLP features. For baseline architecture, we utilize the same architecture described for Wiki-103: a 10-layer decoder-only architecture with eight heads and 128 as model dimensions. We can see that learning MLP feature maps for all the intermediate layers (i.e., layers 2, 4, and so on) is helpful compared to just attending to the final and middle layers. Further, if we remove the MLP features and attend to the original embeddings, we still beat  we see a substantial gain in learning MLP features, though we still beat our baseline architecture by a thin margin. 
\vspace{-0.5cm}
\begin{table}[ht]
  \caption{Comparison Of Model Dimension For wiki103}
	\centering
	\begin{tabular}{|c|c|c|}
		\hline
		Model Dimension & NLL & Accuracy\\\hline
		32  dim Baseline & 5.01  & 23.27\% \\
        32 Attending All Layers & 4.95 & 23.94\%\\
        128 dim Baseline &  3.95 & 31.78\% \\
        128 Attending All Layers  & \textbf{3.86} & \textbf{32.81\%} \\
      
       \hline		
	\end{tabular}
	\label{tab:example}
\end{table}
\vspace{-0.5cm}
\subsection{Scaling with Size} Since the experiments carried out in this paper were at a small academic scale, we see the effect of model dimension for one of the datasets. For this, we conduct two experiments with model dimension 32 and model dimension 128 while retaining the same topology as previously reported. The scaling laws for LLMs are well documented. We only carry this out for the Wiki-103 corpus and report the results in Table 3.

\vspace{-0.3cm}
\subsection{Attention Maps} 
\begin{figure} 
  \centering
  \hspace*{-9.8pt}
  \includegraphics[width=\linewidth]{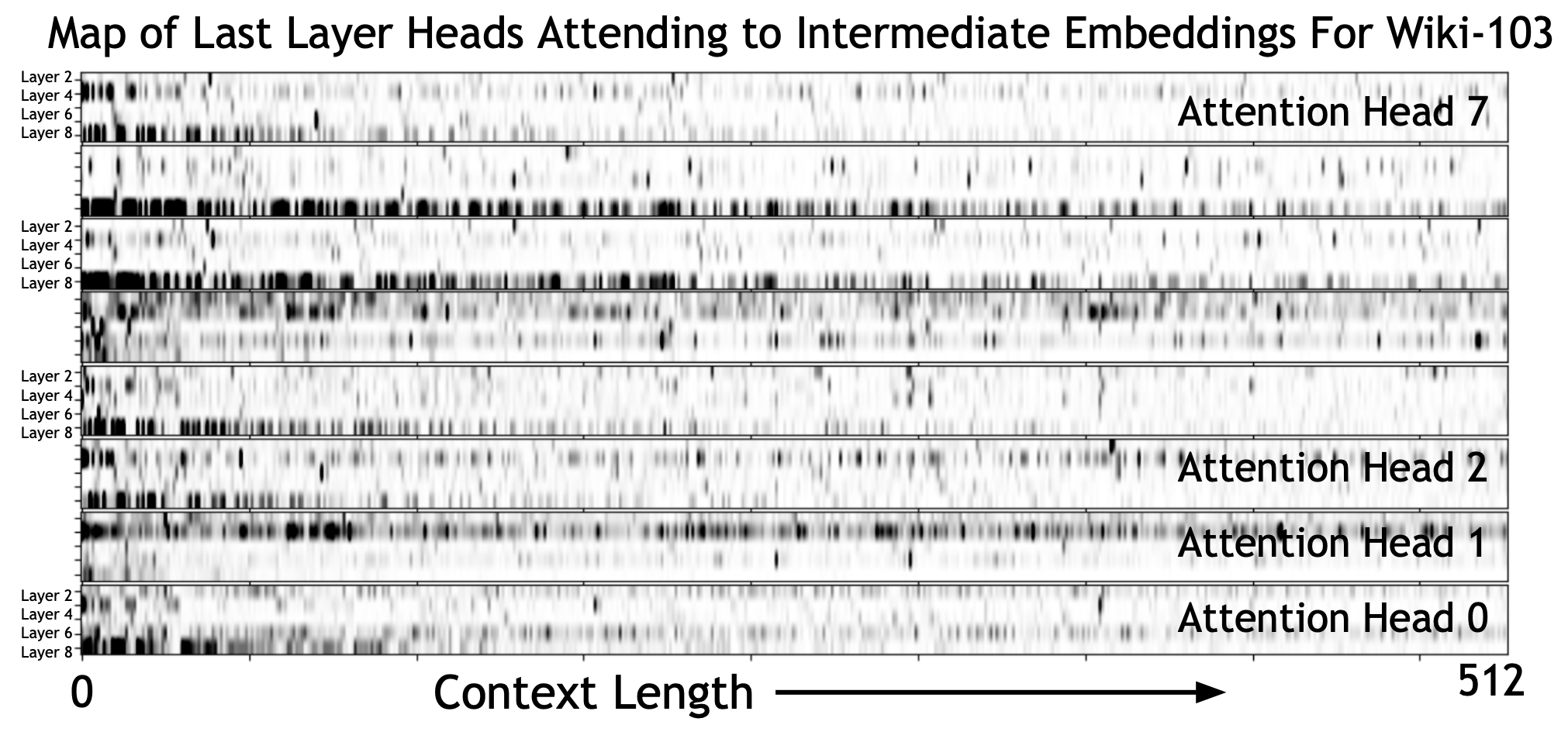}
  \caption{Learned attention maps show how we can adaptively attend to any intermediate layer depending on input.}
  \label{fig:attention_map}
\end{figure}

We would expect the architecture to learn complex dependencies as per the contents of the input tokens. This is because, for tokens where it is simpler to predict the next token, the model need not waste layers to learn a linear or an MLP mapping. It must be able to attend to it directly, bypassing the computations afterward. To show this, we take the attention maps of the last layer that attends to all intermediate representations. We take an input with sequence length 512 from the Wiki-103 dataset and compute the attention maps. As shown in Figure \ref{fig:attention_map}, for every token in the context length, different attention heads give weightage to the learned MLP features from different layers instead of keeping the weights constant.
Further, these change for every token for different attention heads, giving the model much more flexibility than a simple serialized operation. We see in Figure \ref{fig:attention_map} that in almost all of the heads, for initial tokens, most of the weight is given to the final layers. This is intuitive, as in the initial layers, there is minimal amount of context present. Therefore, our architecture tries hard and utilizes deeper layers for prediction. 
\label{sec:intro}

\section{Conclusion And Future Work} We have showcased a method that improves large language model pre-training on small academic-scale architectures by allowing the neural architecture to attend to the intermediate feature representations. We give evidence that the attention map attends to all of the intermediate representations depending on the contents of the input signal. This allows the architecture to develop attention shortcuts as it deems fit to directly attend to utilize shallow representations if necessary, thereby freeing the subsequent layers to carry out heavier computations. Finally, exploring several possible variants of our model will be interesting, for example, combining it with pathways and other kinds of efficient neural architecture.

\bibliographystyle{IEEEbib}
\bibliography{refs}

\begin{thebibliography}{10}

\bibitem{vaswani2017attention}
Ashish Vaswani~et. al,
\newblock ``Attention is all you need,''
\newblock in {\em Advances in neural information processing systems}, 2017, pp.
  5998--6008.

\bibitem{brown2020language}
T.~Brown~et, al,
\newblock ``Language models are few-shot learners,''
\newblock {\em arXiv preprint arXiv:2005.14165}, 2020.

\bibitem{verma2021generative}
Prateek Verma and Chris Chafe,
\newblock ``A generative model for raw audio using transformer architectures,''
\newblock {\em 2021 24th International Conference on Digital Audio Effects
  (DAFx)}, pp. 230--237, 2021.

\bibitem{huang2018music}
Anna Huang~et. al,
\newblock ``Music transformer: Generating music with long-term structure,''
\newblock in {\em International Conference on Learning Representations (ICLR)},
  2019.

\bibitem{verma2020framework}
Prateek Verma and Julius Smith,
\newblock ``A framework for contrastive and generative learning of audio
  representations,''
\newblock {\em arXiv preprint arXiv:2010.11459}, 2020.

\bibitem{borsos2023audiolm}
Zal{\'a}n Borsos~et. al,
\newblock ``Audiolm: a language modeling approach to audio generation,''
\newblock {\em IEEE/ACM Transactions on Audio, Speech, and Language
  Processing}, 2023.

\bibitem{yan2021videogpt}
Wilson Yan, Yunzhi Zhang, Pieter Abbeel, and Aravind Srinivas,
\newblock ``Videogpt: Video generation using vq-vae and transformers,''
\newblock {\em arXiv preprint arXiv:2104.10157}, 2021.

\bibitem{team2023gemini}
Gemini Team,
\newblock ``Gemini: a family of highly capable multimodal models,''
\newblock {\em arXiv:2312.11805}, 2023.

\bibitem{Chameleon}
Team Chameleon,
\newblock ``Chameleon: Mixed-modal early-fusion foundation models,''
\newblock {\em arXiv:2405.09818}, 2024.

\bibitem{hewitt2019structural}
John Hewitt and Christopher~D Manning,
\newblock ``A structural probe for finding syntax in word representations,''
\newblock in {\em Proceedings of the 2019 Conference of the North American
  Chapter of the Association for Computational Linguistics: Human Language
  Technologies, Volume 1 (Long and Short Papers)}, 2019, pp. 4129--4138.

\bibitem{tenney2019bert}
I~Tenney,
\newblock ``Bert rediscovers the classical nlp pipeline,''
\newblock {\em arXiv preprint arXiv:1905.05950}, 2019.

\bibitem{song2020utilizing}
Youwei Song, Jiahai Wang, Zhiwei Liang, Zhiyue Liu, and Tao Jiang,
\newblock ``Utilizing bert intermediate layers for aspect based sentiment
  analysis and natural language inference,''
\newblock {\em arXiv preprint arXiv:2002.04815}, 2020.

\bibitem{oh2022don}
Dongsuk Oh, Yejin Kim, Hodong Lee, H~Howie Huang, and Heuiseok Lim,
\newblock ``Don't judge a language model by its last layer: Contrastive
  learning with layer-wise attention pooling,''
\newblock {\em arXiv preprint arXiv:2209.05972}, 2022.

\bibitem{pasad2023comparative}
Ankita Pasad, Bowen Shi, and Karen Livescu,
\newblock ``Comparative layer-wise analysis of self-supervised speech models,''
\newblock in {\em ICASSP 2023-2023 IEEE International Conference on Acoustics,
  Speech and Signal Processing (ICASSP)}. IEEE, 2023, pp. 1--5.

\bibitem{gong2023whisper}
Yuan Gong~et. al,
\newblock ``Whisper-at: Noise-robust automatic speech recognizers are also
  strong general audio event taggers,''
\newblock {\em arXiv preprint arXiv:2307.03183}, 2023.

\bibitem{hernandez2023linearity}
Evan Hernandez~et. al,
\newblock ``Linearity of relation decoding in transformer language models,''
\newblock {\em ICLR}, 2024.

\bibitem{chowdhery2023palm}
Aakanksha Chowdhery~et. al,
\newblock ``Palm: Scaling language modeling with pathways,''
\newblock {\em Journal of Machine Learning Research}, vol. 24, no. 240, pp.
  1--113, 2023.

\bibitem{panayotov2015librispeech}
Vassil Panayotov, Guoguo Chen, Daniel Povey, and Sanjeev Khudanpur,
\newblock ``Librispeech: an asr corpus based on public domain audio books,''
\newblock in {\em 2015 IEEE international conference on acoustics, speech and
  signal processing (ICASSP)}. IEEE, 2015, pp. 5206--5210.

\bibitem{defossez2022high}
Alexandre D{\'e}fossez~et. al,
\newblock ``High fidelity neural audio compression,''
\newblock {\em arXiv preprint arXiv:2210.13438}, 2022.

\bibitem{wang2023neural}
Chengyi Wang, Sanyuan Chen, Yu~Wu, Ziqiang Zhang, Long Zhou, Shujie Liu, Zhuo
  Chen, Yanqing Liu, Huaming Wang, Jinyu Li, et~al.,
\newblock ``Neural codec language models are zero-shot text to speech
  synthesizers,''
\newblock {\em arXiv preprint arXiv:2301.02111}, 2023.

\bibitem{hawthorne2018enabling}
Curtis et.~al Hawthorne,
\newblock ``Enabling factorized piano music modeling and generation with the
  maestro dataset,''
\newblock in {\em Proceedings of the International Conference on Learning
  Representations (ICLR)}, 2019.

\bibitem{hawthorne2022general}
Curtis Hawthorne~et. al,
\newblock ``General-purpose, long-context autoregressive modeling with
  perceiver ar,''
\newblock in {\em International Conference on Machine Learning}. PMLR, 2022,
  pp. 8535--8558.

\bibitem{al2019character}
Rami Al-Rfou, Dokook Choe, Noah Constant, Mandy Guo, and Llion Jones,
\newblock ``Character-level language modeling with deeper self-attention,''
\newblock in {\em Proceedings of the AAAI conference on artificial
  intelligence}, 2019, vol.~33, pp. 3159--3166.

\bibitem{yu2024megabyte}
Lili Yu, D{\'a}niel Simig, Colin Flaherty, Armen Aghajanyan, Luke Zettlemoyer,
  and Mike Lewis,
\newblock ``Megabyte: Predicting million-byte sequences with multiscale
  transformers,''
\newblock {\em Advances in Neural Information Processing Systems}, vol. 36,
  2024.

\bibitem{kaplan2020scaling}
Jared Kaplan, Sam McCandlish, Tom Henighan, Tom~B Brown, Benjamin Chess, Rewon
  Child, Scott Gray, Alec Radford, Jeffrey Wu, and Dario Amodei,
\newblock ``Scaling laws for neural language models,''
\newblock {\em arXiv preprint arXiv:2001.08361}, 2020.

\end{thebibliography}

\end{document}